# Experimental Demonstration of Array-level Learning with Phase Change Synaptic Devices

S. Burc Eryilmaz[1], Duygu Kuzum[2], Rakesh G. D. Jeyasingh[1], SangBum Kim[3], Matthew BrightSky[3], Chung Lam[3] and H.-S. Philip Wong[1]

[1] Department of Electrical Engineering, Stanford University, Stanford, CA 94305, e-mail: eryilmaz@stanford.edu,
[2] University of Pennsylvania, Philadelphia, PA, [3] IBM Research, T.J. Watson Research Center, Yorktown Heights, NY

## Abstract

The computational performance of the biological brain has long attracted significant interest and has led to inspirations in operating principles, algorithms, and architectures for computing and signal processing. In this work, we focus on hardware implementation of brain-like learning in a brain-inspired architecture. We demonstrate, in hardware, that 2-D crossbar arrays of phase change synaptic devices can achieve associative learning and perform pattern recognition. Device and array-level studies using an experimental 10×10 array of phase change synaptic devices have shown that pattern recognition is robust against synaptic resistance variations and large variations can be tolerated by increasing the number of training iterations. Our measurements show that increase in initial variation from 9 % to 60 % causes required training iterations to increase from 1 to 11.

## I. Introduction

Synaptic electronics is an emerging field of research aiming to build electronic systems that mimic computational energy-efficiency and fault tolerance of biological brain in a compact space [1]. To date, synaptic electronics research has primarily focused on device level implementations of synaptic plasticity and array or chip level simulations [2-4]. Studies that involve on-chip demonstration of brain-like computation with synaptic devices and integration of synaptic devices in a compact, massively parallel architecture remain to be the next big challenge [5]. Recent brain imaging studies suggest that neural connections are in the form of folding 2D sheets of parallel neuronal fibers that cross paths at right angles [6]. The grid-like connectivity of the brain can be best emulated using a 2-D or layered 3-D crossbar array architecture with two terminal synaptic devices (Figure 1). In this work, we experimentally demonstrate that brain-like pattern recognition can be implemented using 2-D crossbar arrays of phase change synaptic devices. First we present individual synaptic device characteristics for the phase change devices in the 10×10 array (section II), and then we explain array level operation and programming scheme and algorithms used for learning (section III). Finally, analysis on resistance variation and its effect on array level learning and pattern completion are discussed (section IV).

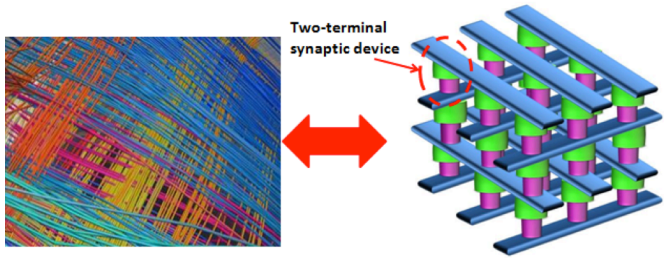

**Figure 1**: Left figure is a DSI (diffusion spectrum imaging) scan showing a fabric-like 3-D grid structure of connections in the monkey brain (Credit: Van Wedeen, M.D., Martinos Center and Dept. of Radiology, Massachusetts General Hospital and Harvard University Medical School) [6]. Right figure is an electronic implementation of 3D crossbar array with synaptic devices inspired from grid-like structure of neuronal fibers.

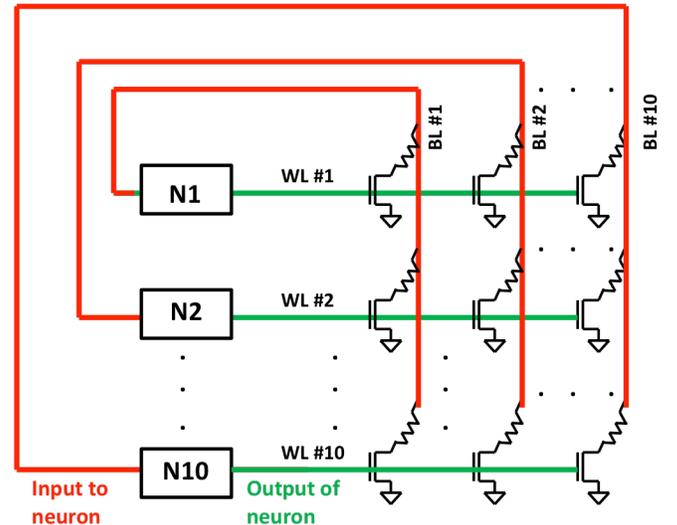

**Figure 2**. Memory array schematic shown as a 10-neuron recurrent neural network. The resistors represent the phase change memory cells (PCM).

## II. Synaptic Array

Nanoscale devices based on phase change memory, PCM ($Ge_2Sb_2Te_5$), have been shown to mimic biological synapses [2-4]. In the experiments, 10-by-10 PCM cell arrays with selection transistors were used (Fig. 2) [7,8]. Ten bitlines (BL) and ten wordlines (WL) are connected to the top electrode of the device and the gate of the selection transistors, respectively. Microscope image of the array and TEM image of a single device [7,8] are shown in Figure 3(left) and (right), respectively. Typical DC switching and pulse switching characteristics of a single device arbitrarily chosen from one of the arrays are shown in Figures 4 and 5, respectively. Set and reset pulses with amplitudes of 1 V and 1.5 V and with (50 ns/300 ns/1 μs) and (20 ns/50 ns/5 ns) rise/width/fall time used at WL node results in a uniform distribution across the array (Fig. 6). Gradual programming to achieve analog storage is implemented by applying multiple pulses to the same PCM cells (Fig. 7). By adjusting pulse amplitudes and widths, it is possible to implement more gradual switching using PCM cells [2,3].

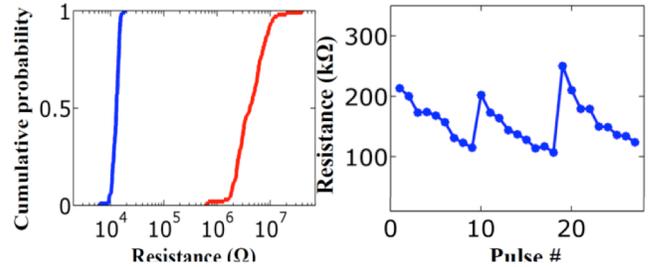

**Figure 6**. Binary resistance distribution of the memory

**Figure 7**. Gradual SET characteristics of a single cell.

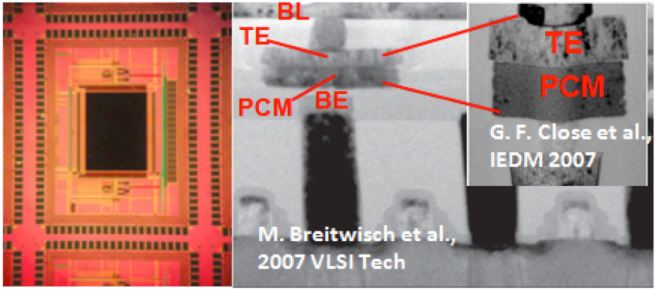

**Figure 3**. Array pad microscope image (left) and single memory cell schematic and TEM image (right) [7,8].

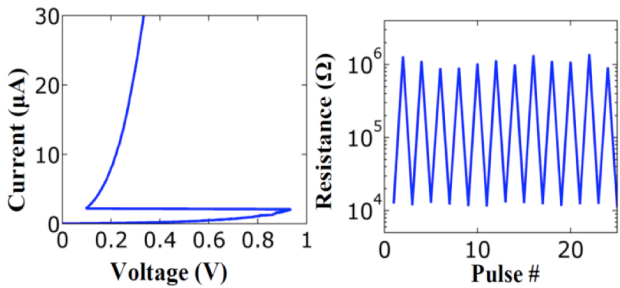

**Figure 4.** Single cell DC switching.

**Figure 5**. Single cell pulse switching between high resistance and low resistance states.

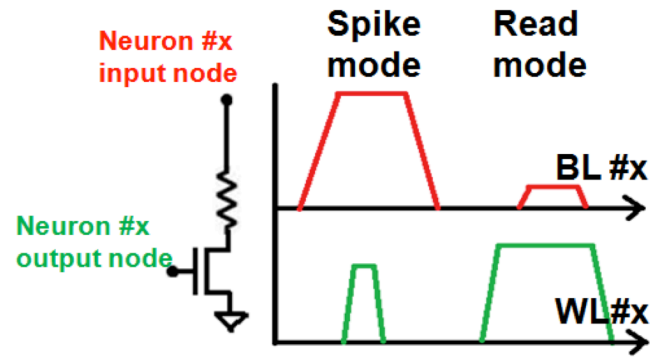

**Figure 8**. Single neuron and its connections to WL and BL for a firing neuron. If a neuron is not firing at an iteration, spike mode signals and read mode WL signals are not applied at that neuron, only BL read pulse is applied.

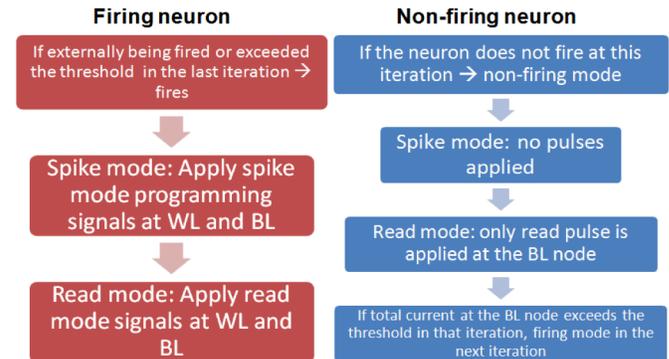

**Figure 9**. Algorithm flowchart for one iteration. Left shows a firing neuron and right shows a non-firing neuron.

## III. Array-level Learning

The crossbar array consists of 100 synaptic devices and 10 recurrently connected neurons (Fig. 2). Integrate-and-fire neurons are implemented by computer control and PCM cells were used as synaptic devices between neurons. The input terminal of each neuron is connected to a BL node, whereas the output terminal of each neuron is connected to a WL node (Fig. 8). In the learning phase, synaptic weights are updated via a simple form of Hebbian plasticity. The synapses between coactive neurons that fire in the same 100 μs time window, get stronger. Otherwise, synaptic weight does not change. At each learning iteration (epoch), if a neuron is not firing, it integrates the current from all the inputs due to the spiking neurons at that iteration. This is achieved by applying a 0.1 V read pulse at input node (BL) of a non-firing neuron. If the total current through the input node of a non-firing neuron is large enough to exceed the threshold, it fires in the next iteration. Firing can also occur due to external stimulation of a neuron. If a neuron is in firing mode at an iteration, it sends a gating pulse to the output node (WL) and a programming pulse to the input node (BL). This network algorithm is explained in Figure 8 and 9. As an illustration of this learning process, we choose two simple patterns to be learned. The two patterns of 10 pixels are shown in Fig. 10 (b) and (c), and the neuron number associated to each pixel is shown in Figure 10(a).

## IV. Resistance Variation

In order to explore the effect of resistance variation on pattern recognition performance, two arrays with high (60 %) and low (9 %) resistance variation are used. Figure 11 shows initial resistance distribution for Array I (high variation) and Array II (low variation). High variation is obtained by initially applying the same RESET pulse to every cell to bring them to the partial RESET state, and low variation is obtained by carefully selecting pulse amplitudes individually for each cell and bringing them to the fully RESET state. The neurons of both synaptic arrays are stimulated with patterns 1 and 2 during the learning epochs. In Figure 11, darker cells show the synapses which get stronger after 11 training epochs. Note that after feeding each input pattern into the network, synapses between the ON neurons get stronger (resistance decreases). Evolution of cell resistance distribution for Array I (Figure 12) explains that the higher the initial resistance variation is, the more training epochs are required for the network to learn the pattern. After the learning phase, the network is presented with an incomplete pattern. Fig. 13 shows neuron input currents for non-firing neurons for Array I and II. For Array II with low initial resistance variation, one training epoch is sufficient for firing neurons (1-4) to recruit neuron 6 and complete the pattern. Input current for non-firing neurons seem to have more variation for array I, resulting in a higher firing threshold requirement and hence more epochs (11 for array I) to recall original pattern (Figure 14). Both arrays with low and high initial resistance variation are able to recall original pattern after ten epochs. Total energy consumed to learn the pattern is measured to be 199 pJ and 8.4 nJ for arrays with low and high variation, respectively. Low energy consumption in Array II is due to smaller number of epochs required and the high resistance regime the devices operate. Resistance variations as high as 60% can be tolerated for pattern completion operation at the expense of more learning iterations and energy consumption.

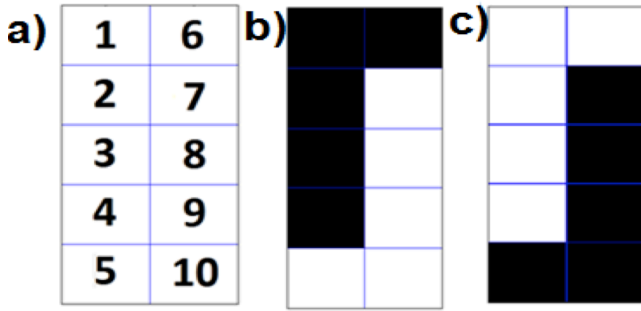

**Figure 10**. Patterns fed into the network a) associated neuron # for each pixel b) pattern1 (neuron #1,2,3,4,6 are ON) c) pattern 2 (neuron #5,7,8,9,10 are ON).

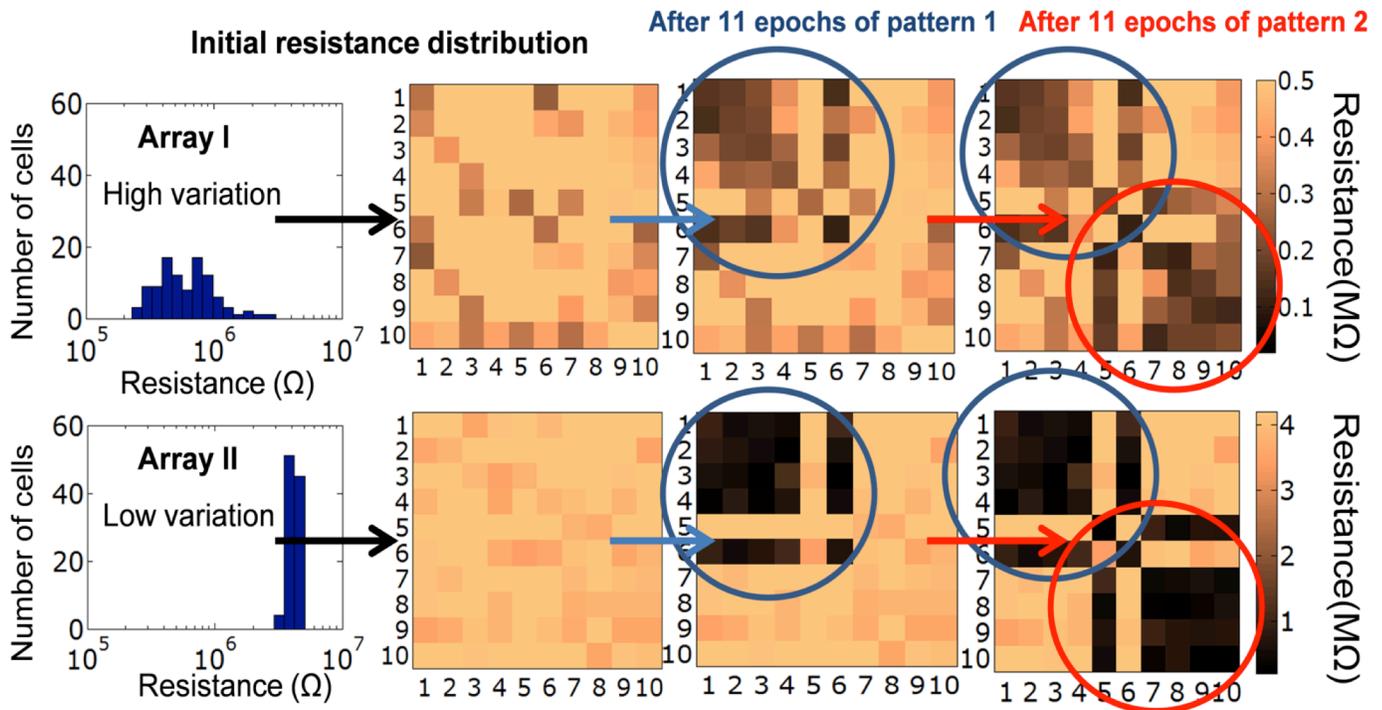

**Figure 11.** Evolution of array resistance distribution after 11 training epochs of pattern 1 and 2, respectively. Experiments are performed on two arrays, one with high initial resistance variation (60 % variation), and the other with low initial resistance variation (9 % variation). Synapses getting stronger by presenting pattern 1 and 2 are shown with blue and red circles, respectively, after training with 11 epochs of each pattern. Synaptic weight contrast in Array II is larger between weak and strong synapses, implying better accuracy of pattern recognition after the same number of epochs.

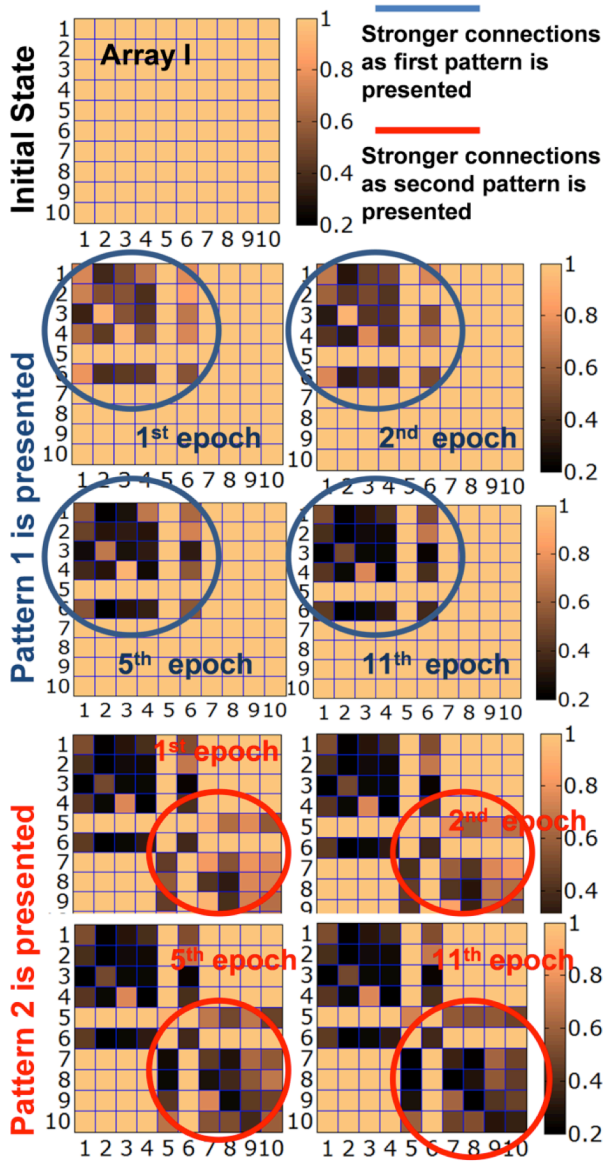

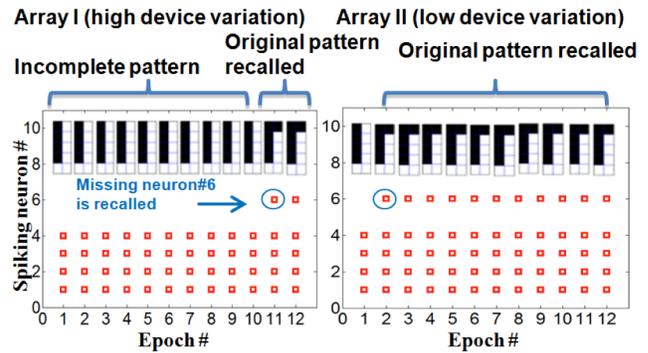

**Figure 14**. Pattern recall. Red squares represent spiking neurons for every epoch. Both arrays with low and high initial resistance distribution were able to recall pattern 1 successfully (note the recall of neuron #6). Array II recalls the original pattern after 1 epochs while Array I needs 11 epochs to recall the original pattern.

## V. Conclusion

We report brain-like learning, in hardware, of a small-scale 10 × 10 crossbar array with synaptic devices. We demonstrated in hardware experiments that synaptic network can implement robust pattern recognition through brain-like learning. Test patterns were shown to be stored and recalled associatively via Hebbian plasticity in a manner similar to the biological brain. Initial resistance variations can be tolerated by adding more training epochs and at the expense of more energy consumption. Demonstration of robust brain-inspired learning in a small-scale synaptic array is a significant step towards building large-scale computation systems with brain-level computational efficiency. Our work inspired from the grid-like connectivity of the brain may lead to more biologically-plausible brain prosthetics in future.

**Figure 12**. Evolution of synaptic weights as pattern 1 and pattern 2 is presented. Resistance values are normalized to initial cell resistances . Synapses between neurons #1,2,3,4,6 and #5,7,8,9,10 gradually get stronger as pattern 1 and 2 are presented, respectively. 11 epochs are needed to get sufficient synaptic weight contrast between weak and strong synapses, since initial resistance variation is high.

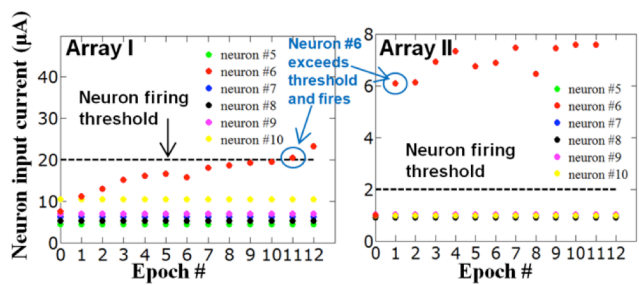

**Figure 13.** Neuron firing threshold is selected to be at least twice the initial state input current of neurons (input current before any training) to prevent undesired pixel recall. Since Array I requires a higher threshold for the same accuracy due to high initial variation, it requires larger number of epochs to recall the original pattern as shown.

## Acknowledgments

This work is supported in part by SONIC, one of six centers of STARnet, a Semiconductor Research Corporation program sponsored by MARCO and DARPA, the Nanoelectronics Research Initiative (NRI) of the Semiconductor Research Corporation (SRC) through the NSF/NRI Supplement to the NSF NSEC Center for Probing the Nanoscale (CPN), and the member companies of the Stanford Non-Volatile Memory Technology Research Initiative (NMTRI).

## References

[1] D. Kuzum, S. Yu, H.-S. Philip Wong, *Nanotechnology*, **24**, 382001 (2013).
[2] D. Kuzum, R. G. D. Jeyasingh, B Lee, H. S. P. Wong, *Nano Letters*, pp. 2179-2186 (2012).
[3] D. Kuzum, R. G. D. Jeyasingh, S. Yu, H.-S. P. Wong, *Trans. Elec. Dev.*, pp. 3489-3494, 2012
[4] M. Suri *et al.*, *IEDM*, 4.4.1-4, 2011

[5] Y. Kaneko, Y. Nishitami, M. Ueda, A. Tsujimura, *Symp. VLSI Tech.*, pp. T238-239, 2013
[6] V.J. Wedeen *et al.*, *Science*, vol. 335, pp. 1628-1634, 2012.
[7] M. Breitwisch *et al. Symp. VLSI Tech.*, pp 100-101, 2007.
[8] G. F. Close *et al.*, *IEDM*, pp.29.5.1-4, 2010.